\documentclass{article}

\usepackage{microtype}
\usepackage{graphicx}
\usepackage{booktabs} 
\usepackage{subcaption}
\usepackage{authblk}
\usepackage[english]{babel}
\usepackage[utf8x]{inputenc}
\usepackage[T1]{fontenc}
\usepackage{amsmath}
\usepackage{dsfont}
\usepackage{notations}
\usepackage{enumitem}
\usepackage[noend]{algorithmic}
\usepackage{hyperref}

\usepackage[accepted]{icml2021}

\icmltitlerunning{\Chambent\ for Online AutoML}

\begin{document}
\twocolumn[
\icmltitle{\Chambent\ for Online AutoML}




\begin{icmlauthorlist}
\icmlauthor{Qingyun Wu}{msr}
\icmlauthor{Chi Wang}{msr}
\icmlauthor{John Langford}{msr}
\icmlauthor{Paul Mineiro}{msr}
\icmlauthor{Marco Rossi}{msr}
\end{icmlauthorlist}

\icmlaffiliation{msr}{Microsoft Research}

\icmlcorrespondingauthor{Qingyun Wu}{qxw5138@psu.edu}
\icmlcorrespondingauthor{John Langford}{jcl@microsoft.com}

\icmlkeywords{Machine Learning, ICML}

\vskip 0.3in
]



\printAffiliationsAndNotice{}  

\begin{abstract}
We propose the \Chambent\ (Champion-Challengers) algorithm for making an \emph{online} choice of hyperparameters in online learning settings.
\Chambent\ handles the process of determining a champion and scheduling a set of `live' challengers over time based on sample complexity bounds. 
It is guaranteed to have sublinear regret after the optimal configuration is added into consideration by an application-dependent oracle based on the champions. Empirically, we show that \Chambent\ provides good performance across a wide array of datasets when optimizing over featurization and hyperparameter decisions.
\end{abstract}

\section{Introduction}

\textbf{Motivation}
Online learning services (for example the Personalizer Azure cognitive service ~\cite{Personalizer}) have a natural need for "autoML" style learning which automatically chooses hyperparameter configurations over some set of possible choices.  The well-studied setting of offline autoML strategies~\cite{Bergstra2015hyperopt,feurer2015efficient,ICLR:li2017hyperband,falkner2018,huang2019efficient,JMLR2019NAS,real2020automl} do not satisfy several natural constraints imposed by the online setting.

\begin{enumerate}[nolistsep]
    \item In online settings, computational constraints are more sharply bounded.  Partly, this is for budgetary concerns and partly this is because any functioning online system must keep up with a constantly growing quantity of data.  Hence, we seek approaches which require at most a constant factor more computation and more generally requires that the online learning algorithm keeps up with the associated data stream.
    \item Instead of having a fixed dataset, online learning settings naturally have a specified data source with unbounded, and sometimes very fast, growth.  For example, it is natural to have datasets growing in the terabytes/day range.  On the other hand, many data sources are also at much lower volumes, so we naturally seek approaches which can handle vastly different scales of data volume.
    \item Instead of evaluating the final quality of the model produced by a learning process, online learning settings evaluate a learning algorithm constantly, implying that an algorithm must be ready to answer queries at all times, and that the relevant performance of the algorithm is the performance at all of those evaluations. 
\end{enumerate}

It is not possible to succeed in online autoML by naively applying an existing offline autoML algorithm on the data collected from an online source.  First, this approach does not address the computational constraint, as direct use of offline autoML is impractical when the dataset is terascale or above.  Operating on subsets of the data would be necessary, but the dramatic potential performance differences in learning algorithms given different dataset sizes suggests that the choice of subset size is critical and data-dependent. Automating that choice is non-trivial in general.  Second, this approach does not address the issue of online evaluation, as offline autoML algorithms are assessed on the quality of the final configuration produced. It is also not possible to succeed in online autoML via naive application of existing regret-minimizing online algorithms~\cite{PLG}: Instantiating a no-regret algorithm over all sets of possible configurations is computationally prohibitive.

How then can we best design an \emph{efficient} online automated machine learning algorithm?

In the online setting there are no natural points for stopping training, evaluating a configuration, and trying the next configuration. If we keep evaluating a fixed set of configurations, other configurations are denied experience, which could lead to linearly increasing total regret. 
Exploration is however delicate because an unknown quantity of data may be necessary for a configuration to exhibit superior performance.  Given this, a simple exploration approach which periodically switches between configurations in a batched epsilon-greedy style may also incur large total regret.  A method that can allocate the limited computational power (at any time point) to learning models while maintaining good online performance (i.e., low regret) and working despite an unknown required example threshold is needed.

\begin{figure}[t]
\begin{center}
\centerline{\includegraphics[width=0.8\columnwidth]{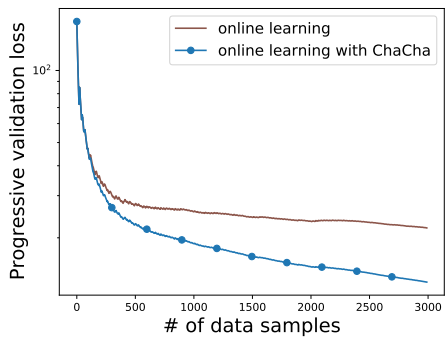}}
\caption{\Chambent\ improves online learning - a demonstration.}
\label{fig:res_demo}
\end{center}
\end{figure}

\subsection{What we do}
In Section~\ref{sec:setting}, we define a new \emph{online} automated machine learning setting with tight computational constraints, consistent with standard online learning settings.  A key element in this setting is a multiplicative constant on the amount of computation available which corresponds to the maximum number of "live" models at any time point. In addition, we assume no model persistence is allowed. The other key element is the availability of a configuration oracle which takes as input a configuration and provides as output a set of possible configurations to try next. This oracle is designed to capture the ability of a domain expert or offline autoML methods to propose natural alternatives which may yield greater performance.

The oracle could propose more configurations than can be simultaneously assessed given the computational budget. How can we efficiently discover a better configuration while respecting the computational budget?  It is critical to allocate computation across the possibilities in a manner which both does not starve potentially best possibilities and which does not waste either computation or experience.  The \Chambent\ algorithm detailed in Section~\ref{sec:method} is designed to do this using an amortized computation approach.  Two other critical issues exist: How can a better configuration be reliably identified? And when evaluating multiple (potentially disagreeing) alternatives how should we make predictions? For the former, we use progressive validation sample complexity bounds while for the latter we identify a model using these bounds and predict according to it, providing benefit from a potentially better configuration before it is proved as such.  

In Section~\ref{sec:theory} we analyze theoretical properties of \Chambent. We prove that as long as the oracle can successfully suggest the optimal configuration, 
\Chambent\ can always achieve a sublinear regret bound even in the worst case. In some ideal cases it can achieve a regret bound that matches the regret bound of the optimal configuration. This sanity-check analysis provides some reassurance that the algorithm behaves reasonably in many settings.

We test the \Chambent\ algorithm on a suite of large regression datasets from OpenML~\cite{Vanschoren2014} for two online autoML tasks. Figure~\ref{fig:res_demo} shows a demonstrative result obtained by \Chambent\ for tuning features interactions choices, eclipsing a widely used online learning algorithm. Further experimentation demonstrates \Chambent\ is consistently near-best amongst plausible alternatives.

\subsection{Related work}
There are many interesting autoML systems designed for the offline setting. A few representative examples from academia are Auto-sklearn~\cite{feurer2015efficient}, Hyperopt~\cite{Bergstra2015hyperopt}, Hyperband~\cite{ICLR:li2017hyperband} and FLAML~\cite{wang2021flaml}. There are also many end-to-end commercial services such as Amazon AWS SageMaker, DataRobot, Google Cloud AutoML Tables, Microsoft AzureML AutoML and H2O Driverless AI. Notable hyperparameter optimization methods include: random search~\cite{Bergstra2012rs,ICLR:li2017hyperband}, Bayesian optimization~\cite{snoek2012practical,Bergstra2015hyperopt,feurer2015efficient,falkner2018}, local search~\cite{koch2018autotune, wu2021cost} and methods that combine local and global search~\cite{eriksson2019scalable,wang2021economical}. In addition, progressively increasing resource and early stopping methods~\cite{ICLR:li2017hyperband,falkner2018,huang2019efficient} are usually found useful when training is expensive.
Despite the extensive study, none of these methods is directly applicable to performing efficient autoML in the online learning setting.

An incremental data allocation strategy, which is conceptually similar to the adaptive resource scheduling strategy in this work, for learner selection is proposed in \cite{sabharwal2016selecting}. However, the method is  designed to function only in the offline batch learning setting with a small fixed set of learners.
Other research directions that share strong synergy with autoML in the online learning literature include  parameter-free online learning~\cite{chaudhuri2009parameter,luo2015achieving,orabona2016coin,foster2017parameter,cutkosky2017stochastic} and online model selection~\cite{sato2001online,muthukumar2019best}. 
Many of these works provide theoretical foundations for online model selection in different learning environments, including both probabilistic and adversarial. However, most of these works only address specific hyper-parameters, such as the learning rate. In addition, these works overlook the sharp constraints on computational power encountered in practice.
\cite{dai2020model} views a model in the context of a bigger system with a target of actively controlling the data collection process for online experiments, which is different from the goal of this work.

\section{Learning Setting} \label{sec:setting}

Several definitions are used throughout. Examples are drawn i.i.d. from a data space $\cX \times \cY$ with $\cX$ the input domain and $\cY$ the output domain. A function $f: \cX \to \cY$ maps input features to an output prediction.
A learning algorithm $A: \cX \times (\cX \times \cY)^* \to \cY$ maps a dataset and a set of input features to a prediction. A loss function $l: \cY \times \cY \to \mathbb{R}$ defines a loss for any output and prediction.   $L_{f} \coloneqq \bbE_{(X,Y)}[l(f(X), Y)]$ denotes the true loss of hypothesis $f$ (under the i.i.d. assumption).  $L_{\cF}^* \coloneqq \min_{f \in \cF} \bbE[l(f(\bx_t), y_t)]$ denotes the best loss achievable using the  best fixed choice of parameters in a function class $\cF$, which contains a set of functions. $f^*$ is the best function given loss function $l$ and the data distribution.  

We consider the following online learning setting: at each interaction $t$, the learner receives a data sample $X$ from the input domain, and then makes a prediction of the sample $A(X, D_t)$ based on knowledge from the historical data samples $D_t$. After making the prediction, the learner receives the true answer $Y$ from the environment as a feedback. Based on the feedback, the learner measures the loss, and updates its prediction model
by some strategy so as to improve predictive performance on future received instances. In such an online learning setting, we want to minimize the cumulative loss $\sum_{t=1}^T L_{A(\cdot, D_t)}$ from the online learner $A$ over the whole interaction horizon $T$ compared to the loss of the best function $f^*$. The gap between them is called `regret' and defined as $ R(T) \coloneqq \sum_{t=1}^T (L_{A(\cdot, D_t)} - L_{f^*}) $.

Similar to the case in offline machine learning, hyperparameters are also prevalent in online learning. A naive choice of the configuration $\tilde c$ may lead to a function class that does not contain the best function. In this case $ R(T) \coloneqq \sum_{t=1}^T (L_{\tilde c,t} - L_{f^*}) \geq \sum_{t=1}^T (L^*_{\cF_{\tilde c}} - L_{f^*}) = \Theta(T)$, i.e., linearly increasing regret is inevitable if the wrong hyperparameter configuration is used. 
We propose online autoML to make \emph{online} choices of the configurations drawn from a configuration space $\cC$. For a particular configuration $c \in \cC$, we denote by $A_{c}$ the corresponding online learner instance and $\cF_c$ the corresponding function class for $A_{c}$. We denote by $D_{t,c}$ the data samples received by $A_c$ up to time $t$. 
We denote \(L_{c,t}\coloneqq L_{A_{c}(\cdot, D_{t,c})}\).
We choose $c_t \in \cC$ for prediction at time \(t\) while operating online autoML. Correspondingly, the regret can be rewritten as $ R(T) \coloneqq \sum_{t=1}^T (L_{c_t,t} - L_{f^*}) $. 

To account for practical online learning scenarios, we target the online setting where (1) a maximum number of $b$ "live" models are allowed to perform online learning at the same time; and (2) no model persistence or offline training is allowed, which means that once we decide to replace a `live' model with a new one, the replaced model can no longer be retrieved. After each model replacement, the new live model needs to learn from scratch, i.e., \(D_{t,c}\) is empty when a model is not live.

\section{Method}
\label{sec:method}

\begin{figure*}[ht]
\begin{center}
\centerline{\includegraphics[width=.8\textwidth]{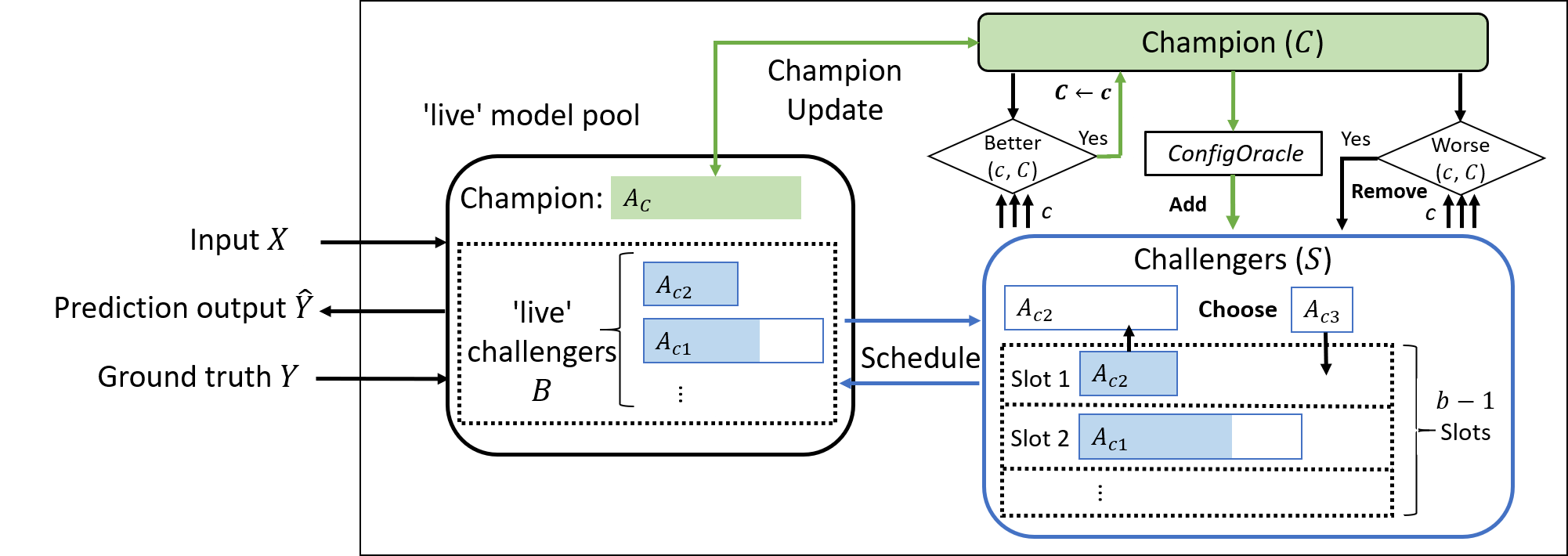}}
\caption{\Chambent\ online autoML framework. The arrows in black are executed at every iteration of online learning. The arrows in green show the operations that are triggered once the \emph{Champion} is updated, and the arrows in blue show the scheduling of `live' \emph{challengers}. Each `live' challenger is represented by a blue rectangle with shaped area, the length of which represents the assigned resource lease. The shaded area of each rectangle reflects how many samples are consumed by each corresponding learner. }
\label{fig:framework}
\end{center}
\end{figure*}

Solving the online autoML problem requires finding a balance between searching over a large number of plausible choices and concentrating the limited computational  budget on a few promising choices such that we do not pay a high `learning price' (regret). \Chambent\ (short for \textbf{Cha}mpion-\textbf{Cha}llengers) is designed to this end with two key ideas: (1) a progressive expansion of the search space according to the online performance of existing configurations; (2) an amortized scheduling of the limited computational resources to configurations under consideration.

To realize these two ideas, we first categorize configurations under consideration in our method into one \emph{Champion}, denoted by $C$, and a set of \emph{Challengers}, denoted by $\cS$.
Intuitively, the \emph{Champion} is the best proven configuration at the concerned time point. The rest of the candidate configurations are considered as \emph{Challengers}. \Chambent\ starts by setting the initial or default configuration, denoted by $c_{init}$ as the champion, and starts with an empty challenger set, i.e, $\cS = \emptyset$ initially. As online learning goes, it (1) updates the champion when necessary and adds more challengers progressively; (2) assigns one of the $b$ slots for `live' models to the champion, and does amortized scheduling of the challengers for the remaining $b-1$ slots when the number of challengers is larger than $b-1$. Under this framework, in the special case where $b=1$, our method degenerates to a vanilla online learning algorithm using the default configuration. In the case where $b>1$, we are able to evaluate more challengers, which gives us a chance to find potentially better configurations. With $b$ `live' models running, \Chambent\ at each iteration selects one of them to do the final prediction. We present the framework of \Chambent\ in Figure~\ref{fig:framework} and provide the pseudocode description in Algorithm~\ref{alg:chambent_framework}. 

\subsection{Progressive search space construction in \Chambent}

\textbf{Generating challengers with champion and \ConfigOracle.}
\Chambent\ assumes the availability of a \ConfigOracle. When provided with a particular input configuration \(c\), \ConfigOracle\ should produce a candidate configuration set that contains at least one configuration that is significantly better than the input configuration \(c\) each time a new configuration is given to it. Such a \ConfigOracle\ is fairly easy to construct in practice. Domain expertise or offline autoML search algorithms can be leveraged to construct such a \ConfigOracle. For example, when the configurations represent feature interaction choices, one typically effective way to construct the oracle is to add pairwise feature interactions as derived features based on the current set of both original and derived features~\cite{luo2019autocross}. With the availability of such a \ConfigOracle, we use a Champion as the `seed' to the \ConfigOracle\ to construct a search space which is then expanded only when a new Champion is identified.

\textbf{A progressive update of the champion using statistical tests.}
\Chambent\ updates the champion when a challenger is proved to be `sufficiently better' than it.  
We use statistical tests with sample complexity bounds to assess the true quality of a configuration and promote new champions accordingly. The central idea of the statistical tests is to use sample complexity bounds and empirical loss to assess the `true' performance $L_{\cF_c}^*$ of the configuration $c$ through a probabilistic lower and upper bound, denoted by $\underline{L}_{c,t}$ and $\overline{L}_{c,t}$ respectively.

Since we consider the online learning setting, we use progressive validation loss~\cite{progressive} as the empirical loss metric.  For a given configuration $c$, we denote by $L^{PV}_{c, t}$ the progressive validation loss of $A_c$ on data sequence $D_{t,c}$.
    \begin{align*}
         L^{PV}_{c, t} \coloneqq \frac{1}{|D_{t,c}|}  \sum_{(X_i,Y_i) \in D_{t,c}} l(A(X_i, D_{t,c}[:i-1]), Y_i) 
    \end{align*}

Without loss of generality, we assume the learning algorithm $A$ can ensure that, for any $\delta>0$,
with probability at least $1-\delta$, $\forall t, |L_{c, t}^{PV} - L_{\cF_{c}}^*| \leq (\comp_{\cF_{c}} \log(|D_{t,c}|/\delta)|D_{t,c}|^{p-1})$ where  
$\comp_{\cF_c}$ denotes a term related to the complexity of the function class $\cF_{c}$, and $0<p<1$ characterizes the estimator error's dependency on the number of data samples. For example, one typical result is $p=\frac{1}{2}, \comp_{\cF} = \sqrt{d}$ for the $d$-dimensional linear regression functions~\cite{10.5555/2621980}.

Based on the bound specified above and an union bound, we can obtain a probabilistic error bound $\epsilon_{c,t}$ which holds for all $t$ and for all $c \in \cS_t$. More formally, with probability at least $1-\delta$, $\forall t, \forall c \in \cS_t$, $|L_{c, t}^{PV} - L_{\cF_{c}}^*| \leq \epsilon_{c,t}$ with
\begin{align} \label{eq:epsilon}
 \epsilon_{c, t}\coloneqq \comp_{\cF_{c}} (\log(|D_{t,c}||\cS_{t}|/\delta))|D_{t,c}|^{p-1})
\end{align}

$ \overline{L}_{c, t} \coloneqq L^{PV}_{c, t}  + \epsilon_{c, t}$ and $\underline{L}_{c, t} \coloneqq L^{PV}_{c, t} - \epsilon_{c, t}$  are thus probabilistic upper and lower bounds of $L^*_{\cF_c}$ for $t \in [T], c \in \cS_{t}$. Based on these probabilistic bounds, we define the following two tests to help distinguish which configuration can lead to better (true) performance.
\begin{equation} \label{eq:test_better}
\BetterThan({c}, {C}, t) 
 \coloneqq \mathds{1}\{  \overline{L}_{{c}, {t}} <  \underline{L}_{{C}, t} -\epsilon_{{C}, t} \}
\end{equation}
\begin{equation} \label{eq:test_worse}
\WorseThan({c}, {C}, t)
\coloneqq
    \mathds{1}\{\underline{L}_{{c}, t}> \overline{L}_{{C}, t} \}
\end{equation}

\Chambent\ eliminates a challenger from consideration once the result of \WorseThan\ test is positive and promotes a challenger to the new champion once the result of a \BetterThan\ test is positive (line 11-15 in Algorithm~\ref{alg:chambent_framework}). When a new champion is promoted, a series of subsequent operations are triggered, including (a) an update of the \Chambent's champion and (b) a call of the \ConfigOracle\, which generates a new set of challengers to be considered. These two operations are reflected in line 8 and line 16-17 of Algorithm~\ref{alg:chambent_framework}
and the green arrows in Figure~\ref{fig:framework}.

Note that when testing whether a challenger $c$ should be promoted into a new champion using the \BetterThan\ test specified in Eq.~\eqref{eq:test_better}, we require the gap between the lower and upper bounds to be at least \(\epsilon_{\Champion,t}\). This ensures that a challenger is promoted into a champion only when it is `sufficiently' better than the old champion, a strategy which avoids the situation where we keep switching champions that are only slightly better than the old ones. That situation is undesirable for two reasons: (a)  
it does not guarantee any lower bound on the loss reduction and thus the true loss between the champion and the true best configuration may remain larger than a constant, which causes a linearly increasing regret in the worst case, and (b) since new challengers are generated and added into consideration, it makes the challenger pool unnecessarily large.

In summary, with the help of \ConfigOracle\ and the statistical tests, \Chambent\ is able to maintain a set of challengers that are not yet `sufficiently better' than the champion but also not significantly `worse' than the champion.

\begin{algorithm}
\caption{\Chambent\ }\label{alg:chambent_framework}
\begin{algorithmic}[1]
\STATE 
\textbf{Inputs:}  Initial configuration $c_{init}$, budget $b$. 

\STATE  \textbf{Initialization:}  $ \Champion \gets c_{init}$;  $\cS \gets \ConfigOracle(\Champion)$; $\Incumbent \gets \Champion$; $\cB = \emptyset$
\FOR{t = 1, ... }
    \STATE \(\rhd\) Observe $X_t$  
    \STATE  $\cB \gets \Scheduler(b, \cB, \cS)$
    \STATE Predict $\hat Y_t \gets A_{\Incumbent}(X_t, D_{t, \Incumbent})$, in which $  {\Incumbent} \gets \argmin_{c \in \cB+\Champion} \overline{L}_{\cF_{c}}$ 
    \STATE  \(\rhd\) Receive $Y_t$ and incur loss $l(\hat Y_t, Y_t)$
\FOR{$c \in \cB+\Champion$}  
\STATE  $\cD_{t+1, c} \gets \cD_{t, c} + (X_t, Y_t)$ and update $A(\cdot, \cD_{t+1, c})$
\ENDFOR
\STATE  
    \(\Champion_{old} \gets \Champion\)
\FOR{$c \in \cS$ }
    \IF{ $ \BetterThan( c, \Champion, t)$} 
       \STATE  $\Champion \gets  c$
     \ENDIF
     \IF{$ \WorseThan(c, \Champion, t)$} 
        \STATE $\cS \gets \cS - c$
     \ENDIF
 \ENDFOR
  \IF{ \(\Champion \neq \Champion_{old}\)}  
  \STATE $\cS \gets \cS + \Frontiers(\Champion)$
  \ENDIF
\ENDFOR

\end{algorithmic}
\end{algorithm}

\subsection{`Live' challenger scheduling}
Now that we have a set of challengers, if the number of `live' model slots is larger than the number of challengers (either because we have a large $b$ or because we have a small $|\cS|$), we can evaluate all the challengers simultaneously. Otherwise we need to perform a careful scheduling. The scheduling problem is challenging since: (1) no model persistence is allowed in the online learning setting so frequent updates of the `live' challengers is costly in terms of learning experience; (2) a blind commitment of resources to particular choices may fail due to those choices yielding poor performance. One principled way to amortize this cost is to use the doubling trick when allocating the sample resource: assign each challenger an initially small lease and successively double the lease over time. We adopt this amortized resource allocation principle and also add a special consideration of the challengers' empirical performance while doing the scheduling. The scheduling step is realized through a \Scheduler function in \Chambent. 

Specifically, the \Scheduler function takes as input  the budget $b$, the current `live' challenger set $\cB$,  the candidate set $\cS$, and provides as output a new set of live challengers (which can have overlap with the input $\cB$). It is designed to eventually provide any configuration with any necessary threshold of examples required for a regret guarantee. 
We call this resource threshold a \emph{resource lease} denoted by $\overline{n}_{c}$. 
Initially every configuration is assigned a particular minimum resource lease \(\overline{n}_c=n_{min}\) (for example $n_{min} =5\times \# features$). When a configuration has been trained with \(\overline{n}_c\) examples (line 6 in Algorithm~\ref{alg:ScheduleTrials}), i.e., reaches its assigned resource lease, we double this resource lease (line 7 in Algorithm~\ref{alg:ScheduleTrials}). 

To avoid starving a challenger under consideration indefinitely, we remove the challenger which just reached its assigned resource lease, from the `live' challenger pool and add the challenger with the minimum resource lease into the `live' challenger pool (line 9 and line 11-13 of Algorithm~\ref{alg:ScheduleTrials} and Algorithm~\ref{alg:fs_selector}). In addition, to avoid throwing away valuable experience for a promising challenger, we adopt a more conservative exploration: We replace a `live' challenger which reaches its assigned resource lease only if it is not among the top performing (according to loss upper bound) `live' challengers (line 8 in Algorithm~\ref{alg:ScheduleTrials}).
In other words, we use half of the compute resources to exploit the candidates that have good performance for now, and another half to explore alternatives that may have better performance if given more resources. 

With the $b$ `live' models running, at each interaction, \Chambent\ selects one of them to make the prediction (line 6 of Algorithm~\ref{alg:chambent_framework}) following the structural risk minimization principle~\cite{vapnik2013nature}.

\begin{algorithm} 
\caption{$\Scheduler(b, \cB, \cS)$} \label{alg:ScheduleTrials}
\begin{algorithmic}[1] 

\STATE \textbf{Notions}: 
$\overline{n}_{c}$ denotes the resource lease assigned to for $c$;
$n_{c,t}$ denotes the resource consumed by $A_c$ by time $t$, e.g., $n_{c,t} = |D_{t,c}|$

\FOR{$c \in \cB$ } 
\IF{$c \notin \cS$ and $|\cS| > b$ }  
\STATE $\cB \gets \cB - c$ 
\ENDIF 
\ENDFOR 

\FOR{$c \in \cB $}
    \IF{$n_{c,t} \ge \overline{n}_{c} $}
    \STATE $ \overline{n}_{c} \gets 2 \overline{n}_{c}$ 
        \IF{$\overline{L}_{c,t} > \text{median}(\{\overline{L}_{c,t} \}_{c \in \cB})$, and $|\cS| > b$} 
       \STATE $\cB \gets \cB - c$ 
        \ENDIF 
    \ENDIF
\ENDFOR
\WHILE{$|\cB| < b-1$}
    \STATE $c \gets \Choose(\cS \setminus \cB)$ 
    \STATE $\cB \gets \cB + c$ 
    \STATE $D_{t+1,c} \gets \emptyset$ \COMMENT{Assuming no persistence of models}
\ENDWHILE
\STATE \textbf{return} \(\cB\)
\end{algorithmic}
\end{algorithm}

\begin{algorithm} 
\caption{$\Choose(\cS)$} \label{alg:fs_selector}
\begin{algorithmic}[1] 

\STATE $\cC^{\text{Pending}} \gets \{c \in \cS: \overline{n}_{c} \text{~has not been set}\} $ 
\IF{$|\cC^{\text{Pending}}| \neq 0$} 

\STATE  $c \gets \text{Random}(\cC^{\text{Pending}})$ and set $\overline{n}_{c} \gets n_{\min}$
\STATE \textbf{return} c
\ENDIF 
\STATE \textbf{return} $\argmin_{c \in \cS } \overline{n}_{c}$ 
\end{algorithmic}
\end{algorithm}

\section{Theory}
\label{sec:theory}
For the convenience of analysis, we split the total time horizon $T$ into $M+1$ phases, the index of which ranges from $0$ to $M$. Phase $0$ starts from $t=1$ and a new phase starts once a new champion is found.  We denote by $t_{m}$ and $t_{m+1}-1$ the starting and end time index of phase $m$ respectively. Thus $N_m \coloneqq t_{m+1}-t_m$ is the length of phase $m$. We denote by $\cS_{m}$ the set of candidate configurations and $\Champion_m$ the champion configuration at phase $m$. 
\begin{equation*}\label{eq:phase_split}
\underbrace{1, 2, \!\cdots\!, \!\cdots\! t_{1}-1}_{ \text{phase 0}}, \cdots, \underbrace{t_m, \!\cdots\!, t_{m+1}-1}_{  \text{phase } m  }, \!\cdots,\!\underbrace{t_M,  \!\cdots\!, T}_{ \text{phase } M}
\end{equation*}

We denote by $c^*$ the optimal configuration, and thus $L_{f^*} = L^*_{\cF_{c^*}}$. To make meaningful analysis, we assume $c^*$ is added in the candidate configuration pool after the \ConfigOracle\ is called a constant number of times.

\begin{lemma} \label{lemma:tests_implication}
With $\epsilon_{c,t}$ being set as in Eq.~\ref{eq:epsilon} and $0< \delta <1$, $\forall m \in [M]$,

\textbf{Claim 1.} $\forall c \in \cS_{m}$, if $ L^*_{\cF_c} - L^*_{\cF_{\Champion_m}} + 2 \epsilon_{c, {t}} + 3  \epsilon_{\Champion_m, {t}} < 0$, with probability at least $1-\delta$, $c$ can pass the \BetterThan\ test described in Eq.~\eqref{eq:test_better} when compared with ${\Champion_m}$ at time \(t\).

\textbf{Claim 2.} When the \BetterThan\ test is positive, with probability at least $1-\delta$, $L^*_{\cF_{\Champion_{m}}}- L^*_{\cF_{\Champion_{m+1}}} > \epsilon_{\Champion_m, t_{m+1}}$. 

\textbf{Claim 3.} If $L^*_{\cF_c} < L_{\cF_{\Champion_m}}^*$, with probability at least $1-\delta$, $c$ will not pass the \WorseThan\ test when compared with ${\Champion_m}$. 
\end{lemma}

\begin{proposition} \label{prop:champion_loss_gap}
With a base learning algorithm having a sample complexity based error bound of $\comp_{\cF_{c}} \log(|D_{t,c}|/\delta)|D_{t,c}|^{p-1}$ and $\epsilon_{c,t}$ being set as in Eq.~\eqref{eq:epsilon}, with high probability, \Chambent\ can obtain,
\begin{align} \label{eq:prop_loss_gap_bound}
   & \sum_{m=0}^{M} \sum_{t=t_{m}}^{t_{m+1}-1} (L^*_{\cF_{\Champion_m}} - L^*_{\cF_{c^*}}) \\ \nonumber
   = &  \tilde O\big(  \max_{m\in[M]}\frac{|\cS_m|}{b}\comp_{\cF_{c^*}} T^p +  \sum_{m=0}^M \comp_{\cF_{\Champion_m}} N_m^{p} \big)
   \\ \nonumber
   = & 
    \tilde O\big( \max_{m\in[M]} \frac{|\cS_m|}{b}\comp_{\cF_{c^*}} T^p  +  \comp_{\cF_{\Champion_m}}  T^{\frac{1}{2-p}} \big)
\end{align}
\end{proposition}
\textbf{Proof intuition of Proposition~\ref{prop:champion_loss_gap}.} 
The proof is mainly based on the first two claims of Lemma~\ref{lemma:tests_implication}. Claim 1 of Lemma~\ref{lemma:tests_implication} ensures that, for all phases $m \in [M]$, the gap between `true' loss of the champion and the best configuration in the pool is with high probability upper bounded by $\epsilon_{c^*, t}$ and $\epsilon_{C_m, t}$. Since $\epsilon_{c,t}$ shrinks with the increase of data samples for $c$, and our scheduling strategy ensures that the optimal configuration $c^*$ receives at least $\frac{b}{4\max_{m \in M } |\cS_m|} T$ data samples, we can obtain an upper bound of $ \tilde O( \max_{m\in[M]}\frac{|\cS_m|}{b}\comp_{\cF_{c^*}} T^p )$ for the summation term on $\epsilon_{c^*,t}$ over $T$ iterations. The summation over $\epsilon_{\Champion_m, t}$, i.e., $\sum_{m=0}^{M} \sum_{t=t_{m}}^{t_{m+1}-1}\epsilon_{\Champion_m, t}$ is delicate because we must account for the case where the champion is  updated frequently, which makes \(M\) and \(\epsilon_{\Champion_m,t}\) large. Fortunately, we designed the \BetterThan\ test such that we switch to a new champion only when it is sufficiently better than the old one which ensures an upper bound on the number of switches, i.e., $M$. Specifically, with Claim 2 of Lemma~\ref{lemma:tests_implication}, and the fact that $\sum_{m=0}^M N_m = T$, we can prove $M = O(T^{\frac{1-p}{2-p}})$. With that we are able to prove $\sum_{m=0}^{M} \sum_{t=t_{m}}^{t_{m+1}-1}\epsilon_{C_m, t} = \tilde O(\sum_{m=0}^{M}  N_m^p) =  \tilde O(T^{\frac{1}{2-p}})$, which contributes to the second term of the bound in Eq.~\eqref{eq:prop_loss_gap_bound}.
 
\begin{remark}[Special cases of Proposition~\ref{prop:champion_loss_gap}]
Proposition~\ref{prop:champion_loss_gap} provides an upper bound of $\sum_{m=0}^{M} \sum_{t=t_{m}}^{t_{m+1}-1} (L^*_{\cF_{\Champion_m}}-L_{\cF_{c^*}}^*) $ in the most general case. It provides guarantees even for the worst cases where the algorithm needs to pay high switching costs, i.e., with a large number of phases  $M$. Such bad cases happen when the algorithm frequently finds new Champions from newly added candidates, whose accumulated sample number is still small and thus $|D_{t_{m+1}, \Champion_m}|$ can only be lower bounded by 1, $\forall m \in [M]$.  In the special case where $|D_{t_{m+1}, \Champion_m}|\geq z |D_{t_{m}, \Champion_{m-1}}|$, in which $z$ is a constant larger than 1 for all $m$, we have $\sum_{m=0}^M N_m^p \log N_m = O(T^p \log T)$, which reduces the upper bound in Proposition~\ref{prop:champion_loss_gap} to $\tilde O\big( T^p \big)$. In this special case, the bound matches the order of regret (w.r.t. $T$) for using the optimal configuration. 
\end{remark}

Without further assumptions,  Proposition~\ref{prop:champion_loss_gap} is tight.
For any small $\eta>0$, we can construct a sequence of $N_m=\Theta(m^q)$, where \(q=\frac{1-p}{\frac{1}{2-p}-p-\eta}-1=\frac{(1-p)(2-p)}{1-(2-p)(p+\eta)}-1>\frac{2-p}{1-p}-1=\frac{1}{1-p}\). Such a sequence satisfies Eq.~\eqref{eq:N_sum_upper_bound}, and makes \(\sum_{m=0}^M N_m^p\log N_m=\Omega(T^{\frac{1}{2-p}-\eta})\).

\begin{theorem} \label{thm:final_regret_bound}
Under the same conditions as specified in Proposition~\ref{prop:champion_loss_gap}, \Chambent\ can achieve the following regret bound, $ \sum_{t=1}^{T} (L_{{\Incumbent_t}, {t}} -L^*_{\cF_{c^*}})  =  \tilde O\big( \max_{m\in[M]} \frac{|\cS_m|}{b}\comp_{\cF_{c^*}} T^p  + \comp_{\cF_{\Champion_m}}  T^{\frac{1}{2-p}} \big)$.
\end{theorem}

\textbf{Proof intuition of Theorem~\ref{thm:final_regret_bound}.} We decompose the regret in the following way $\sum_{t=1}^{T} (L_{{\Incumbent_t},t} -L^*_{\cF_{c^*}}) = \sum_{m=0}^{M} \sum_{t=t_m}^{t_{m+1}-1} (L_{{\Incumbent_t},t} -  L^*_{\Champion_m} ) + \sum_{m=0}^{M}  \sum_{t=t_m}^{t_{m+1}-1} ( L^*_{\Champion_m} - L^*_{\cF_{c^*}})$, the bound of which is quite straightforward to prove once the conclusion in Proposition~\ref{prop:champion_loss_gap} is obtained. 


\section{Empirical Evaluation}
\label{sec:exp}

\ChaCha\ is general enough to handle the online tuning of various types of hyperparameters of online learning algorithms in a flexible and extensible way. It can be tailored to satisfy customized needs through customized implementations of the \ConfigOracle. 

\textbf{Out-of-the-box \ChaCha\ with a default \ConfigOracle} We provide a default implementation of the \ConfigOracle\ such that \ChaCha\ can be used as an out-of-the-box solution for online autoML. This default \ConfigOracle\ leverages an existing offline hyperparameter optimization method~\cite{wu2021cost} for numerical and categorical hyperparameters suggestion. In addition to these two types of typically concerned hyperparameters, we further extend the \ConfigOracle\ by including another important type of hyperparameter, whose configurations space can be expressed as a set of polynomial expansions generated from a fixed set of singletons. 
One typical example of this type of hyperparameter is the choice of feature interactions or feature crossing on tabular data~\cite{luo2019autocross}, where raw features (or groups of raw features) are interacted, e.g., through cross product, to generate an additional set of more informative features. In this example, the set of raw feature units are the set of singletons, and each resulting feature set with additional feature interactions can be treated as one polynomial expansion based on the original singletons. 
This type of featurization related hyperparameter is of particular importance in machine learning practice: it has been well recognized that the performance
of machine learning methods, including both offline batch learning and online learning, depends greatly on the quality of features~\cite{domingos2012few, NIPS2014_8f1d4362}. 
The tuning of this type of hyperparameter is notoriously challenging because of the doubly-exponentially large search space: for a problem with $m$ raw singletons, each interaction (up to a maximum order of $m$) involves any subset of the $m$ raw singletons, implying a ($\sum_{i=0}^{m} {m \choose i} - m -1 = 2^{m}-m-1$)-dimensional binary vector space defining the set of possible interactions.  Since any subset of the polynomial expansions is allowed, there are $2^{2^{m}-m-1}$ possible choices. For this type of hyperparameter, only heuristic search approaches such as Beam search~\cite{medress1977speech}, are available even in the offline learning setting.
Inspired by these offline greedy search approaches and domain knowledge obtained from online learning practice, we realize the \ConfigOracle\ for this type of hyperparameter in the following greedy approach by default in \ChaCha: given the input configuration, \ConfigOracle\ generates all configurations that have one additional second order interaction on the input configuration. For example, given a input configuration $\Champion = \{e_1, e_2, e_3\}$, the \ConfigOracle(\(\Champion\))\ outputs the following set of configurations $\{\{e_1, e_2, e_3, e_1 e_2 \}, \{e_1, e_2, e_3, e_1 e_3\}, \{ e_1, e_2, e_3, e_2 e_3 \} \}$. Under this design, when provided with an input configuration with $k$ groups of features, the \ConfigOracle\ generates a candidate set with  $\frac{k(k-1)}{2}$ configurations.

\begin{figure*} 
\centering 
\begin{subfigure}{0.45\textwidth}
\includegraphics[width=\textwidth]{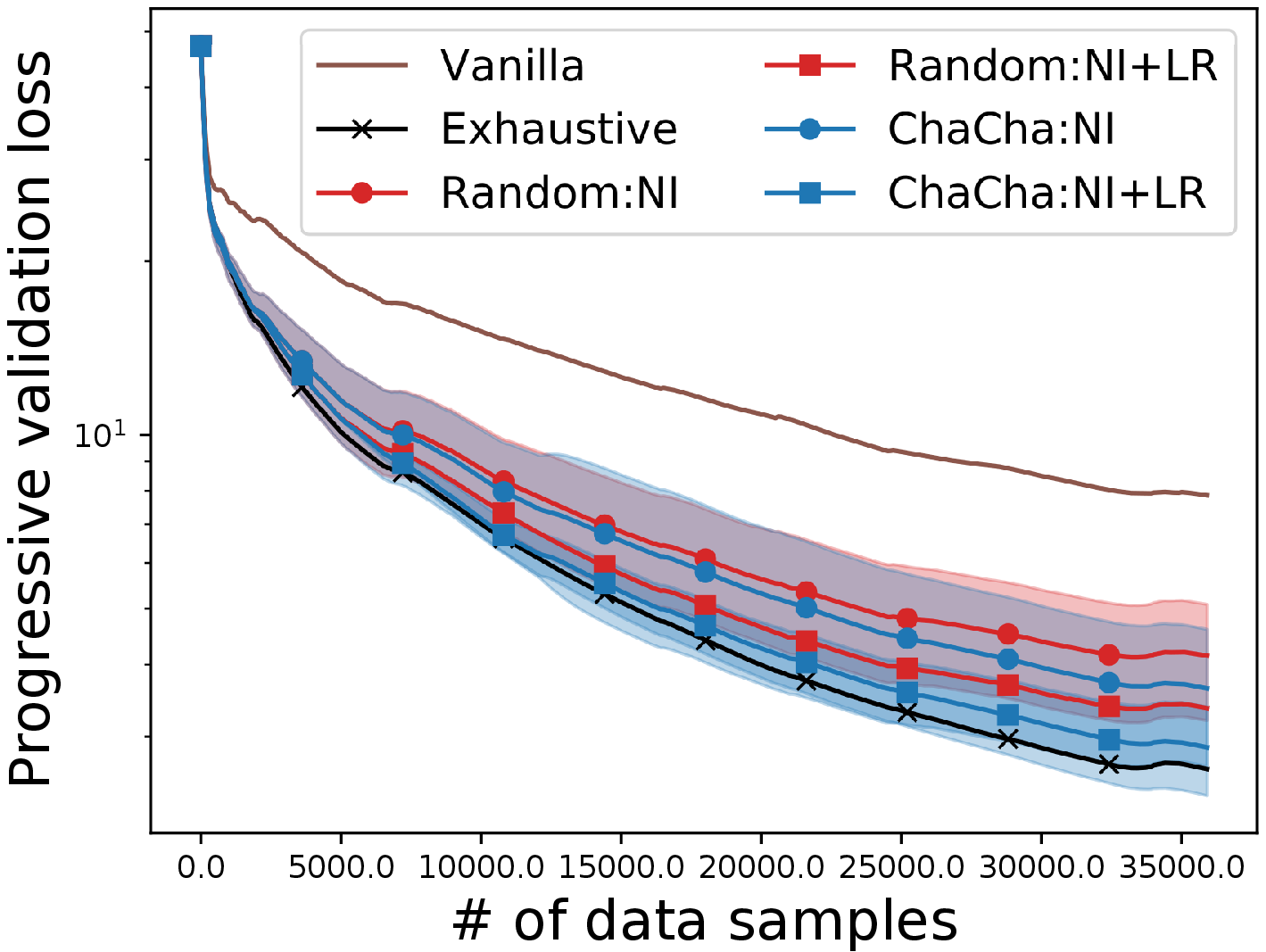}%
\caption{Progressive validation loss over time on dataset \# 41506} 
\end{subfigure}\hfill
\begin{subfigure}{0.45\textwidth}
\includegraphics[width=\textwidth]{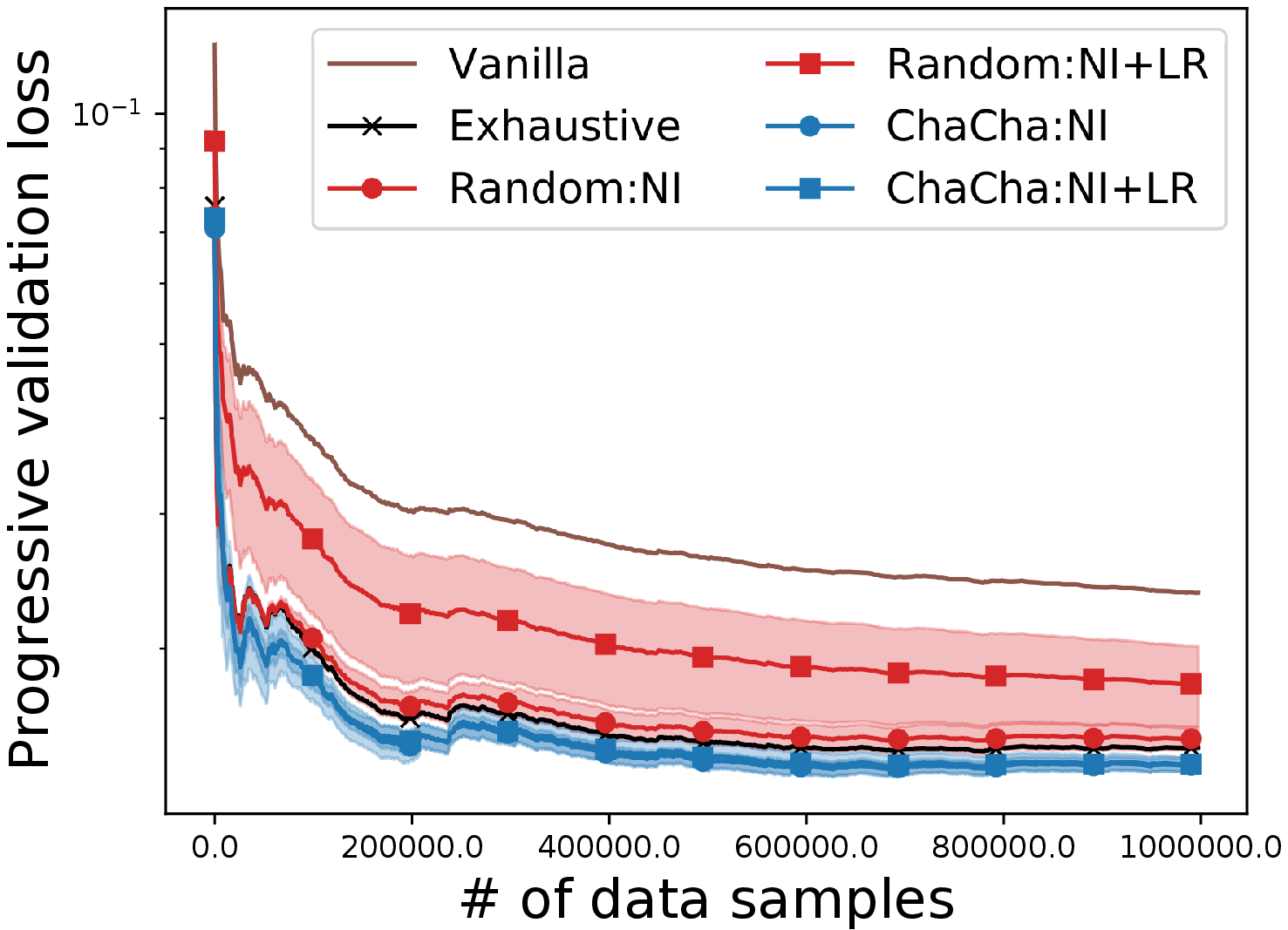}%
\caption{Progressive validation loss over time on dataset \# 5648} 
\end{subfigure}\hfill

\caption{Real time progressive validation loss on a small and large dataset. The shaded area shows the standard derivation of the loss over the 5 runs. The suffixes `:NI' and `:NI+LR' with \ChaCha\ and \RandomInit\ denote the tuning task of namespace interactions, and both namespace Interactions and learning rate respectively. This legend syntax applies to all figures in this paper.}
\label{fig:progressive_validation_loss}
\end{figure*} 

\subsection{AutoML tasks}

\textbf{Online Learning with Vowpal Wabbit.}
Our evaluation is performed using Vowpal Wabbit\footnote{\url{https://vowpalwabbit.org}} (VW), which is an open-source online machine learning library. Users can tune various hyperparameters of online learning algorithms in VW depending on their needs, for example, namespaces interactions, learning rate, l1 regularization and etc.

\textbf{Target hyperparameters to tune.} We perform evaluation in two tuning scenarios to demonstrate both the effectiveness and the flexibility of \ChaCha, including the tuning of namespaces interactions (a namespace is essentially a group of features), and the tuning of both namespace interactions and learning rate. 
The automatic tuning of namespace interactions is of significant importance in VW because (a) it can greatly affect the performance; (b) the challenge exists no matter what online learning algorithms one use; (c) it is not well handled by any of the parameter-free online learning algorithms developed recently. 
For these two tuning tasks, we use a default implementation of the \ConfigOracle\ described earlier in this section. We use the default configuration in VW as the the initial configuration $c_{init}$: no feature interactions, and the learning rate is 0.5.  

We use the VW default learning algorithm (which uses a variant of online gradient descent) as the base learner. We perform the main evaluation under the constraint that a maximum of 5 `live' learners are allowed, i.e., $b=5$. 

\textbf{Baselines and Comparators.} No existing autoML algorithm is designed to handle the online learning scenario, so we compare our method with several natural alternatives.

- \RandomInit: Run learners built on the initial configuration and $(b-1)$ randomly selected configurations from the first batch of candidate configurations generated by the same \ConfigOracle\ used in \ChaCha. 

- \Exhaustive: Exhaust all the configurations generated from \ConfigOracle($c_{init}$) ignoring the computational limit on the number of `live' models. This should provide better performance than \ChaCha\ on the first batch of configurations generated from the \ConfigOracle\ with the initial configuration since it removes the computational limits of \ChaCha.  It is possible for \ChaCha\ to perform better despite a smaller computational budget by  moving beyond the initial set of configurations.

- \Naive: Run the learner built on the initial configuration using $c_{init}$ without further namespace interactions.

Note that \Naive\ and \Exhaustive\ are comparators to better understand the performance of the proposed method, rather than baselines in the context of online autoML with the same computational constraints. 

\textbf{Datasets.}  We evaluate our method on a set of large scale (\# of instance: 10K to 1M) regression datasets from OpenML (in total 40). On these datasets, we group the raw features into up to 10 namespaces. Among the 40 large scale openml datasets, 25 do not allow superior performance to \Naive\ amongst the configurations given by \ConfigOracle. We therefore focus on the 15 `meaningful' datasets where \Exhaustive\ is better than \Naive. 

\subsection{Results}

\begin{figure*} 
\centering 
\begin{subfigure}{0.5\textwidth}
\includegraphics[width=\textwidth]{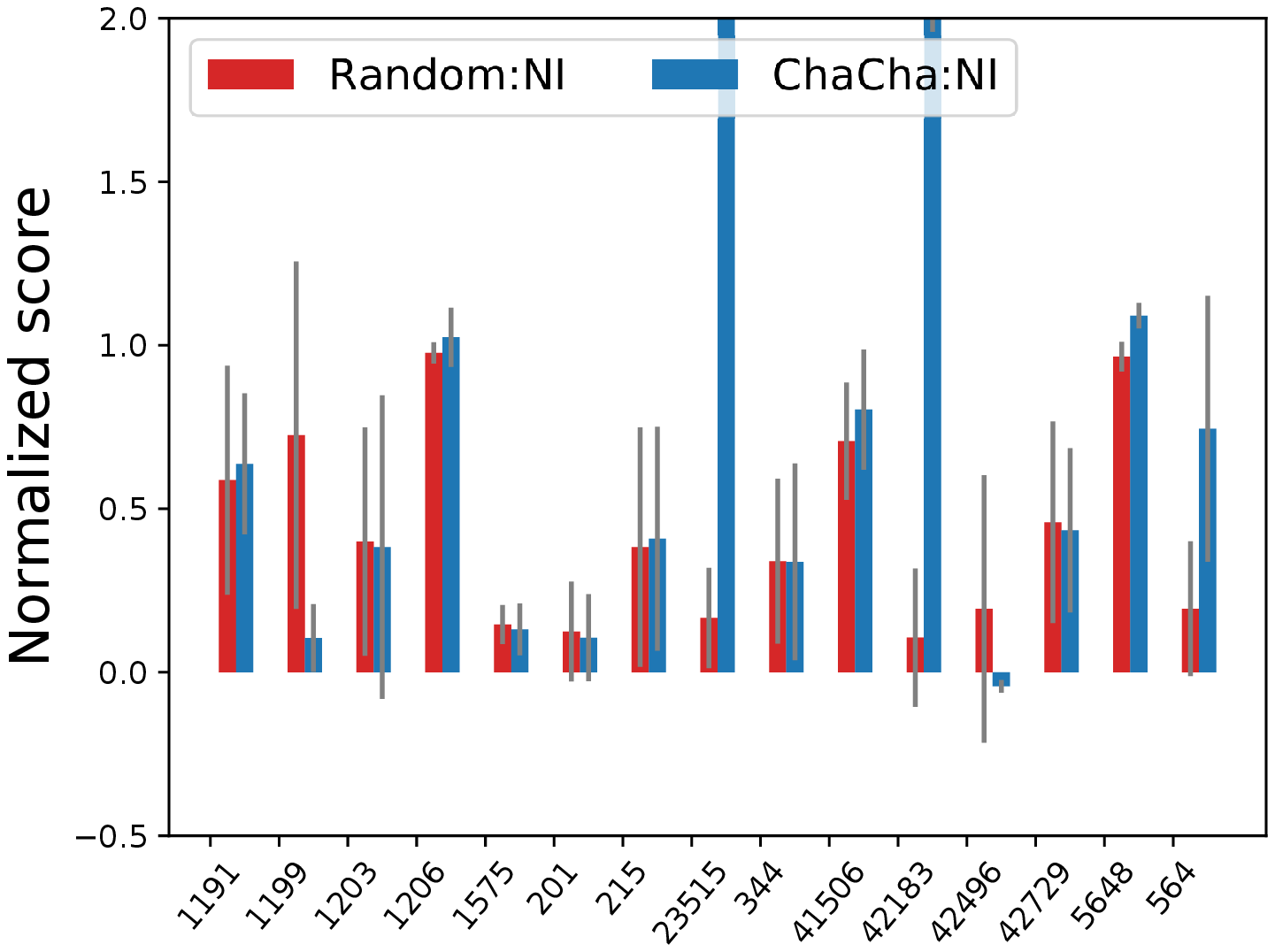}%
\caption{Namespace interactions tuning} 
\end{subfigure}\hfill
\begin{subfigure}{0.5\textwidth}
\includegraphics[width=\textwidth]{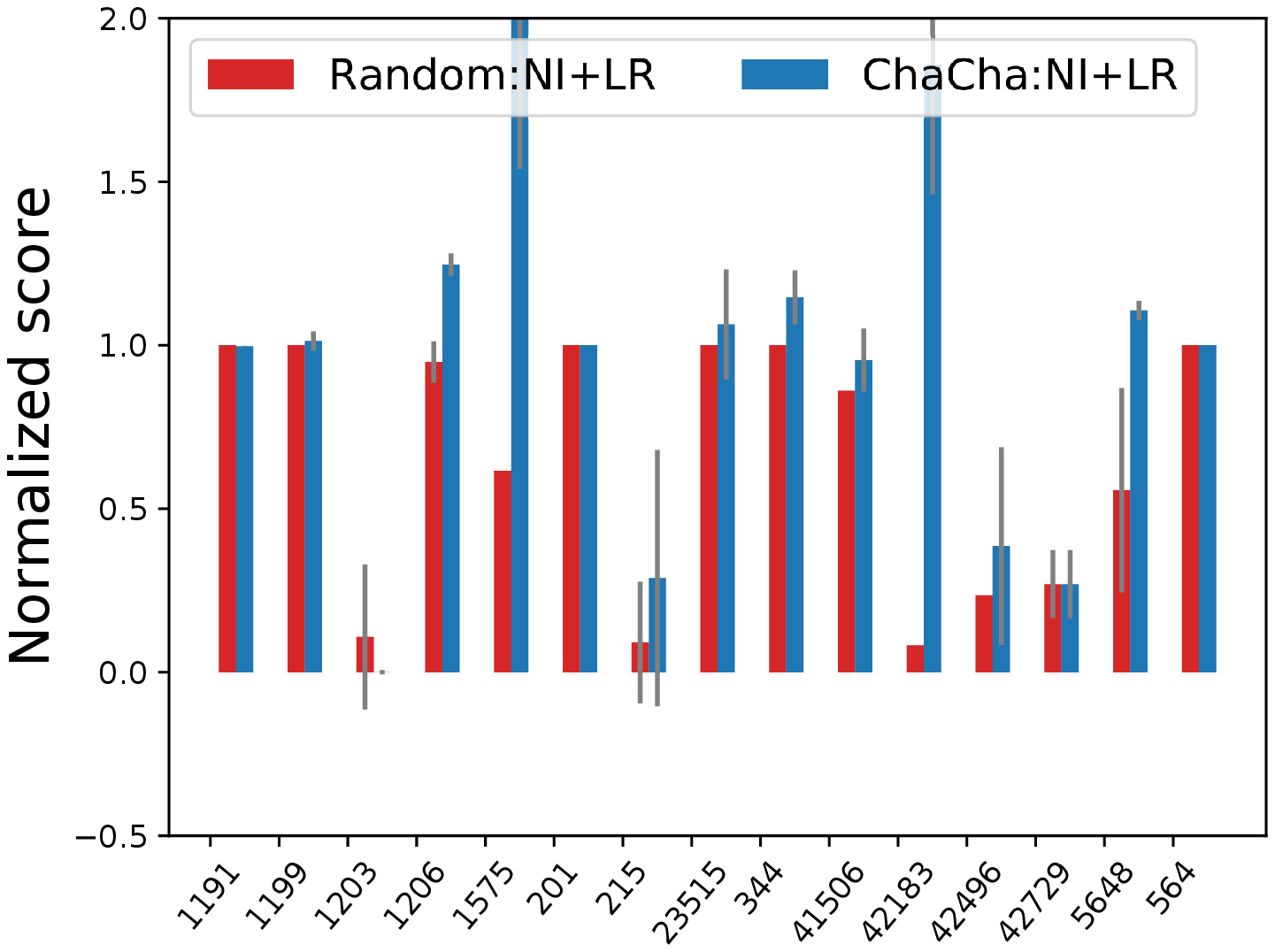}%
\caption{Namespace interactions and learning rate tuning}%
\end{subfigure}\hfill%
\caption{Results on openml datasets for two tuning tasks with error-bars showing the standard derivation over the 5 runs.}
\label{fig:agg_oml}
\end{figure*}

\textbf{Performance on OpenML datasets.} 
Figure~\ref{fig:progressive_validation_loss} shows typical results in terms of the  progressive validation loss (mean square error) on two of the openml datasets. \ChaCha\ is able to outperform all the competitors except \Exhaustive. To show the aggregated results over all the openml datasets, we normalized the progressive validation loss of a particular algorithm using the following formula: $\text{Score(alg)} \coloneqq \frac{L^{PV}(\text{\Naive})  - L^{PV}(\text{alg})   }{L^{PV}(\text{\Naive}) - L^{PV}(\Exhaustive)}$. By this definition, $\text{Score(\Naive)} =0$ and $\text{Score(\Exhaustive)} =1$. The score is undefined when $\text{Score(\Exhaustive)} = \text{Score(\Naive)}$, so we only report results on the `meaningful' datasets where the difference is nonzero. As the serving order of configurations from the candidate configuration pool is randomized, for all the experiments, we run each method 5 times with different settings of random seed (except \Naive\ and \Exhaustive\ as they are not affected by the random seeds) and report the aggregated results. Figure~\ref{fig:agg_oml}(a) and Figure~\ref{fig:agg_oml}(b) show the final normalized scores for two tuning tasks on the 15 datasets after running for up to 100K data samples (or the maximum number of data samples if it is smaller than 100K). We include the final normalized score on datasets with larger than 100K samples in the appendix.
Results in Figure~\ref{fig:agg_oml} show that \ChaCha\ has significantly better performance comparing to \RandomInit\ over half of the 15 `meaningful' datasets. 


\begin{figure}[ht]
\vskip -0.2in
\begin{center}
\centerline{\includegraphics[width=0.9\columnwidth]{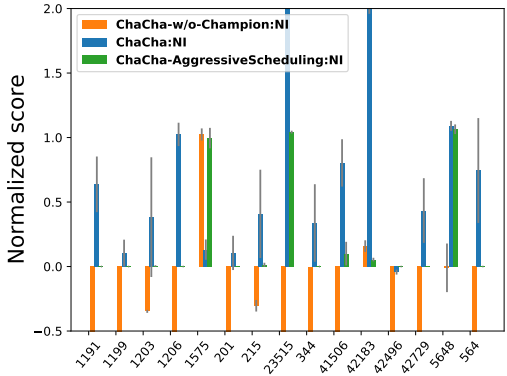}}
\caption{Ablations of \ChaCha\ in namespace interactions tuning.}
\label{fig:ablation}
\end{center}
\vskip -0.1in
\end{figure}

\textbf{Analysis and ablation.} We now investigate several important components in \ChaCha\ to better understand the effectiveness of it. 

\textbf{(1) Search space construction in \ChaCha.} \ChaCha\ uses the champion and \ConfigOracle\ to gradually expand the search space, which leads to a provable improvement on the overall quality of configurations while keeping the size still manageable. The method \RandomInit\ and \Exhaustive\ use a straightforward way to take advantage of the \ConfigOracle: they consider the configurations generated by the \ConfigOracle\ based on the initial configuration. \RandomInit\ works reasonably well if the configuration pool contains many configurations better than the initial configuration. However as expected, it has very large variance. In addition, it is not able to further improve beyond \Exhaustive. The fact that in many cases, \ChaCha\ has better performance than \Exhaustive\ indicates that there is indeed a need for expansion of the search space. In addition to the performance boost comparing to \Exhaustive\ on almost half of the datasets, \ChaCha\ shows almost no performance degeneration comparing to both \Naive\ and \RandomInit. 

\textbf{(2) The scheduling of `live' models.}
Another important component of \ChaCha\ is the scheduling of `live' models.
\ChaCha\ always keeps the champion `live' and schedules $b-1$ `live' challengers in an amortized manner. When scheduling the challengers, to avoid throwing away useful experiences due to the `live' challengers swapping, \ChaCha\ keeps the top-performing half of the `live' challengers running. Results in Figure~\ref{fig:ablation} show the two variants of \ChaCha\ (for the tuning of namespace interactions) without these two designs. 
\ChaCha-AggressiveScheduling denotes the variant of \ChaCha\ in which we do the scheduling purely based on the doubling resource lease without regard for their empirical performance. \ChaCha-w/o-Champion is a variant of \ChaCha\ where we further do not intentionally keep the champion running. The bad results of \ChaCha-w/o-Champion shows the necessity of keeping a learner with sufficiently large training samples `live'. The comparison between \ChaCha-AggressiveScheduling and \ChaCha\ shows the benefit of balancing the long-term and short-term utility of configuration exploration.

We include more evaluation details, the setting of \ChaCha\, and additional results in the appendix. 

\section{Conclusion}
In this work, we propose a novel solution \Chambent\ for Online AutoML. \Chambent\ is a first of its kind autoML solution that can operate in an online learning manner. 
It respects the sharp computational constraints and optimizes for good online learning performance, i.e., cumulative regret. The sharp computational constraint and the need for learning solutions that have a good guarantee on the cumulative regret are unique properties of online learning and not addressed by any existing autoML solution. \Chambent\ is theoretically sound and has robust good performance in practice. 
\Chambent\ provides a flexible and extensible framework for online autoML, which makes many promising future work about online autoML possible. For example, it is worth studying how to make \Chambent\ handle online learning scenarios with bandit feedback. 

\section*{Software and Data}
Our method is open-sourced in the AutoML Libriary FLAML\footnote{\url{https://github.com/microsoft/FLAML/tree/main/flaml/onlineml}}. Please find a demonstration of usage in this notebook\footnote{\url{https://github.com/microsoft/FLAML/blob/main/notebook/flaml_autovw.ipynb}}. All the datasets are publicly available in OpenML\footnote{\url{https://www.openml.org/search?type=data}}.  
\section*{Acknowledgements}
The authors would like to thank Akshay Krishnamurthy for the discussions about this work and the anonymous reviewers for their revision suggestions.

\bibliographystyle{icml2021}
\bibliography{references}
 \newpage
\appendix
\onecolumn
\icmltitle{ \Chambent\ for Online AutoML (Supplementary)}

\section{Evaluation details} \label{sec:appendix_exp}
\subsection{Datasets}
The datasets are obtained from OpenML according to the following criterion: (1) the number of instances in the dataset is larger than 10K; (2) no missing value; (3) regression dataset; (4) the dataset is still active and downloadable. The final list of datasets that satisfy the aforementioned criterion is [573, 1201, 1195, 344, 1192, 201, 216, 41065, 42731, 4545, 42688, 1196, 23515, 1206, 
1193, 42721, 42571, 42713, 537, 42724, 41540, 4549, 296, 574, 218, 5648, 215, 41539, 1199, 1203, 1191, 564, 
1208, 42183, 42225, 42728, 42705, 42729, 42496, 41506].  We converted the original OpenML dataset into VW required format\footnote{\url{https://github.com/VowpalWabbit/vowpal_wabbit/wiki/Input-format}} following the instructions. We sequentially group the raw features of each dataset into up to 10 namespaces\footnote{\url{https://github.com/VowpalWabbit/vowpal_wabbit/wiki/Namespaces}}. Log-transformation on the target variable is performed if the largest value on the target variable is larger than 100 (for datasets whose target variable has negative values, we first shift the value to make it all positive and then do the log-transformation). 

\subsection{Detailed settings of \Chambent\ and baselines}

The two compared methods \RandomInit\ and \Exhaustive\ use the same method as \Chambent\ when selecting one model from the `live' model pool to make the final prediction at each iteration. The minimum resource lease in \Chambent\ is set to be $5 \times$ (dimensionality of the raw features) in all of our experimental evaluations. 
To ensure the empirical loss of the online learning models is bounded, we use the `clipped mean absolute error' as the empirical performance proxy in \Chambent: we keep track of the minimum value, denoted by $\underline y_{t}$, and the maximum value, denoted by $\bar y_{t}$, of the target variable according to observations received up to time $t$. When calculating the mean absolute error for models in \Chambent, we map our prediction of the target variable $\hat y_t$ into this range in the following way: $\min\{\max\{ \underline y_t, \hat y_t\}, \bar y_t\}$. By doing so, we ensure the mean absolute error is always bounded by $\bar y_t - \underline y_t$. Note that this revision is only performed in the update of empirical loss proxy in \Chambent, the final output of \Chambent\ is not clipped. 
For $\epsilon_{c,t}$, we use sample complexity bounds for linear functions. More specifically, $\epsilon_{c,t}$ is set to be $a\sqrt{\frac{d_{c}\log (|D_{c,t}||S_{m_t}|/\delta)}{|D_{c,t}|}}$, in which $d_{c}$ is the dimensionality of the feature induced by namespace configuration $c$, $a$ is constant related to the bound of the loss and is set to be $0.05*(\bar y_t - \underline y_t)$, and $\delta$ is set as $0.1$.

\subsection{Additional Results}
\begin{figure*} [ht]
\centering 
\begin{subfigure}{0.5\textwidth}
\includegraphics[width=\textwidth]{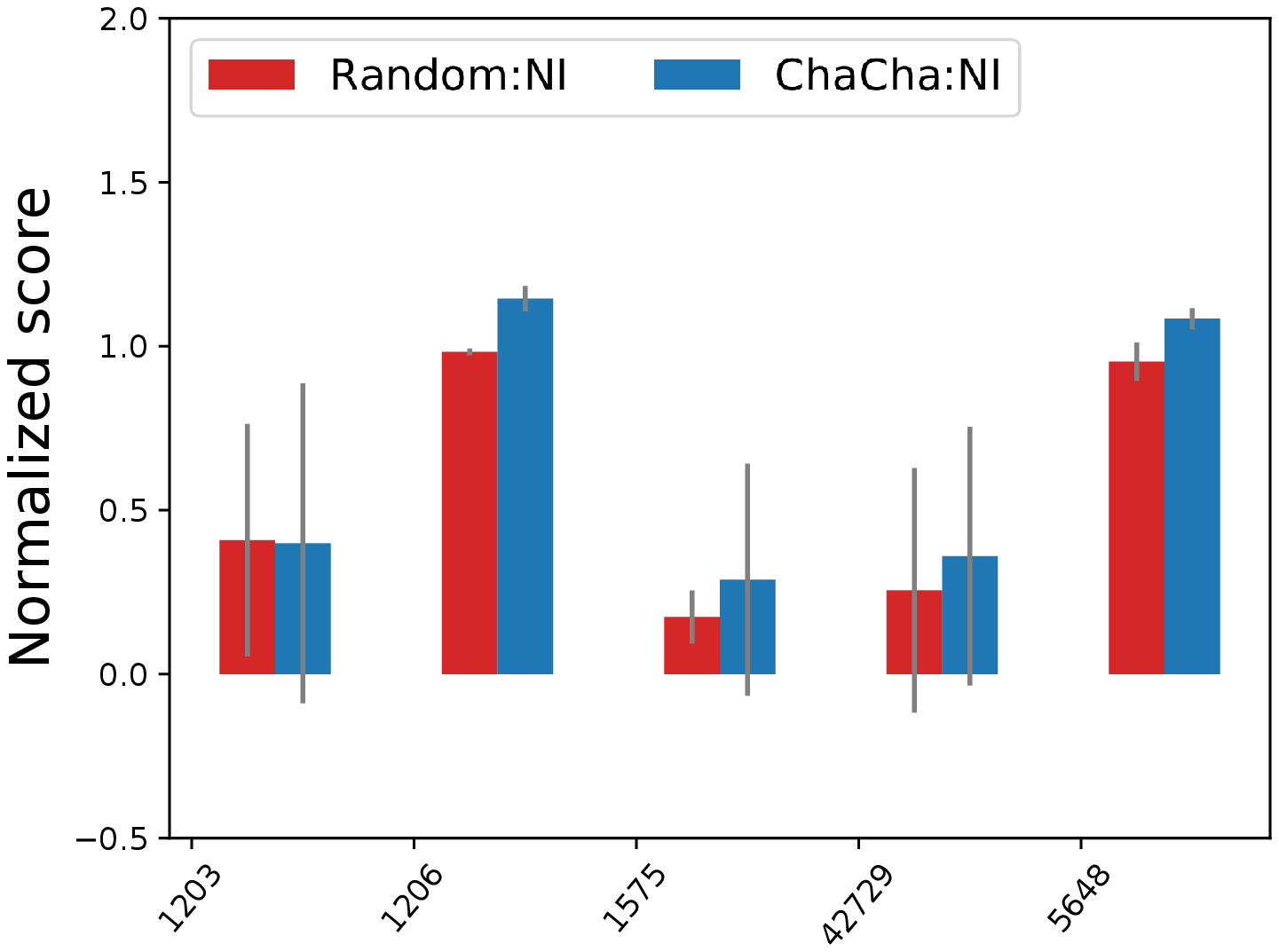}%
\caption{Namespace interactions tuning} 
\end{subfigure}\hfill
\begin{subfigure}{0.5\textwidth}
\includegraphics[width=\textwidth]{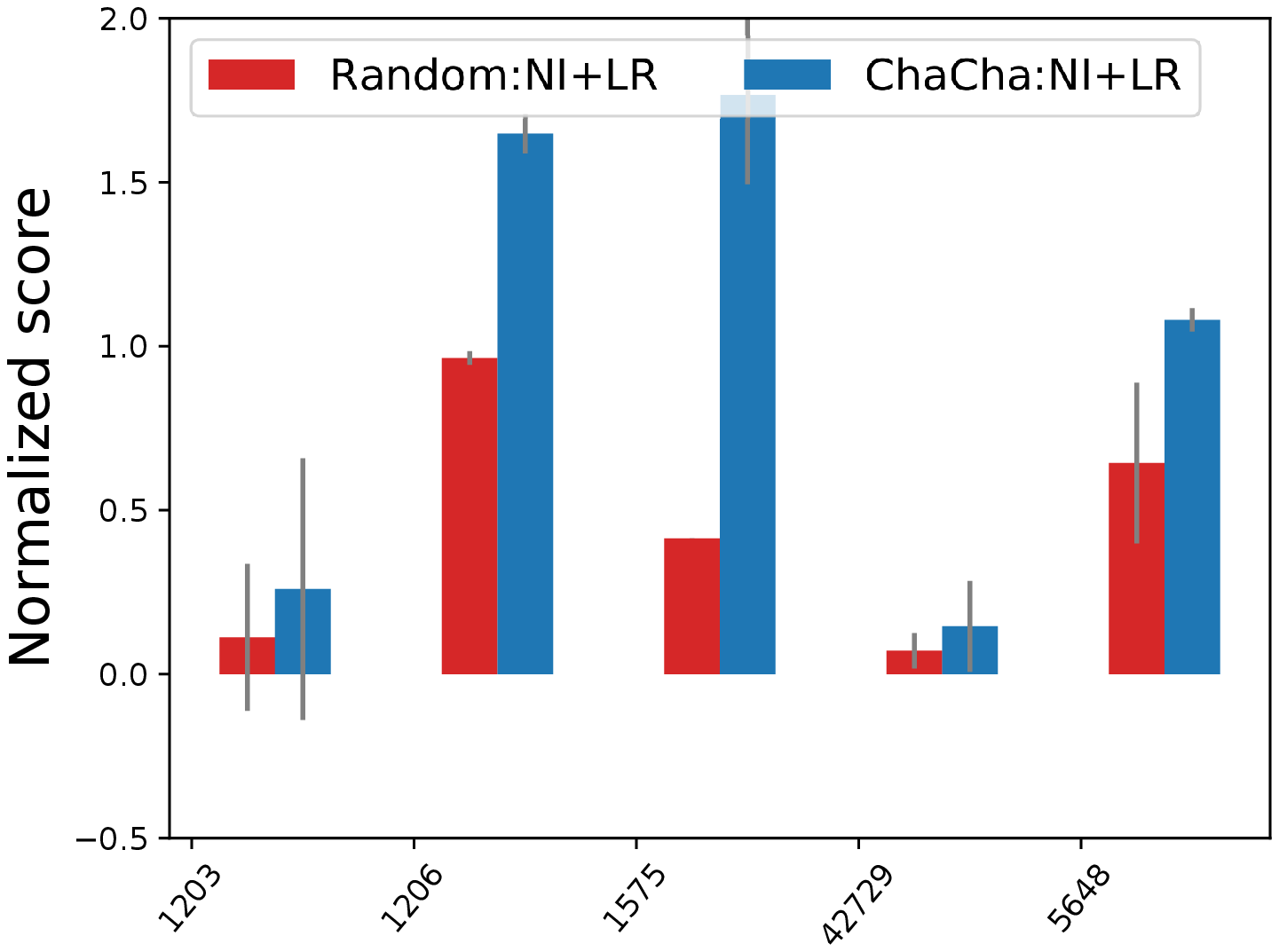}%
\caption{Namespace interactions and learning rate tuning}%
\end{subfigure}\hfill%
\caption{Normalized scores after a larger number of data samples (up to 1M). The normalized scores of \Chambent\ on dataset 1206 for both tuning scenarios, and dataset 1575 and 42729 for namespace interactions tuning are better than those reported in Figure~\ref{fig:agg_oml}, which indicates that \Chambent\ may achieve even larger gain as the increase of data samples.} 
\label{fig:agg_oml_large}
\end{figure*}

We now provide additional results for the cases where all the methods are run for a larger number of data samples. We compare the results on the three largest datasets with up to 1M data samples in Figure~\ref{fig:agg_oml_large}. The result shows the consistent advantage of \Chambent\ under large data volumes. In addition, since several bars in Figure~\ref{fig:agg_oml} and Figure~\ref{fig:ablation} are cropped, we include the actual numbers of the normalized scores in Table~\ref{tab:scores} and Table~\ref{tab:score_ns+lr} for completeness.

\begin{figure*} 
\centering 
\begin{subfigure}{0.5\textwidth}
\includegraphics[width=\textwidth]{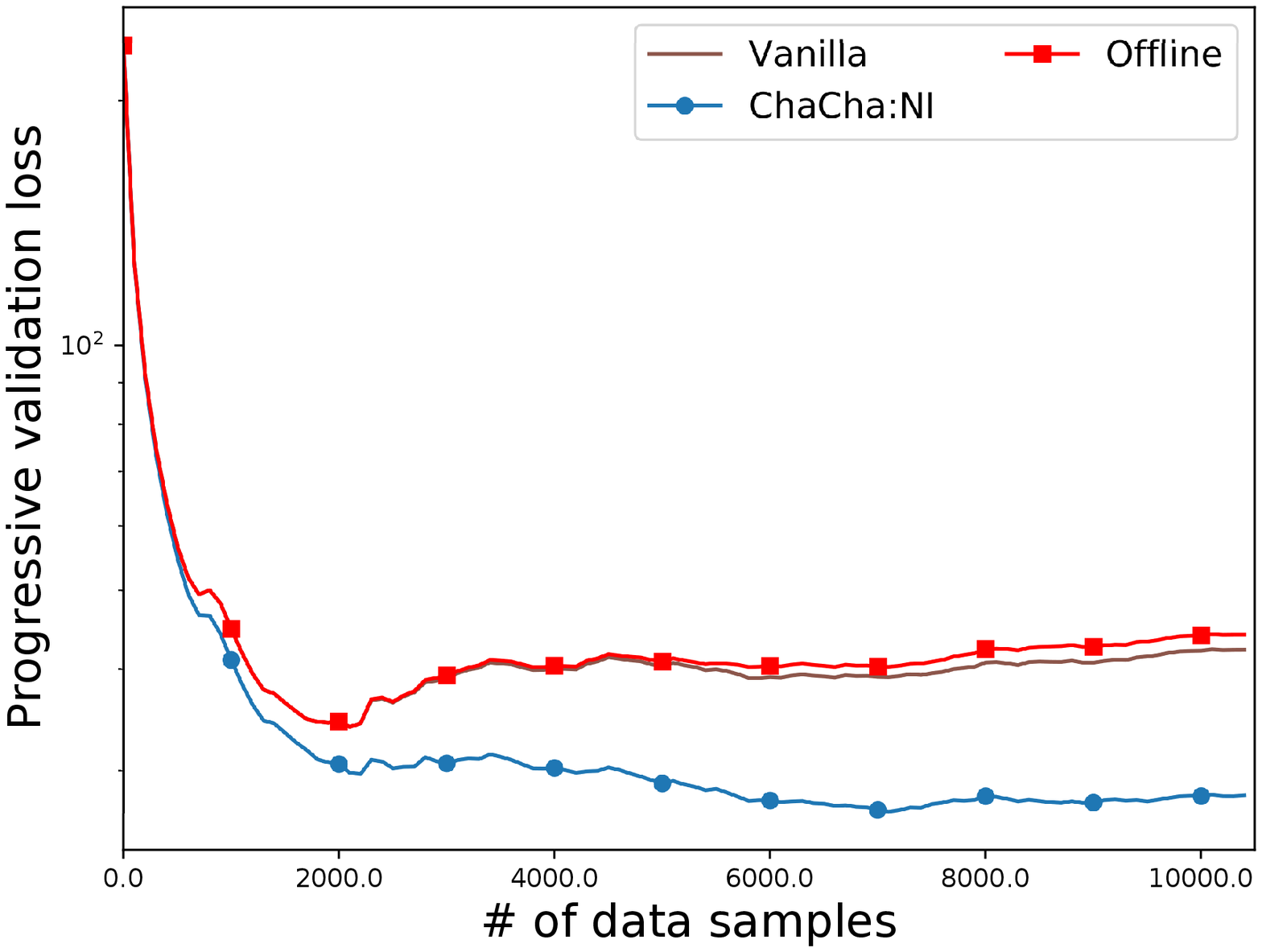}%
\end{subfigure}\hfill
\begin{subfigure}{0.5\textwidth}
\includegraphics[width=\textwidth]{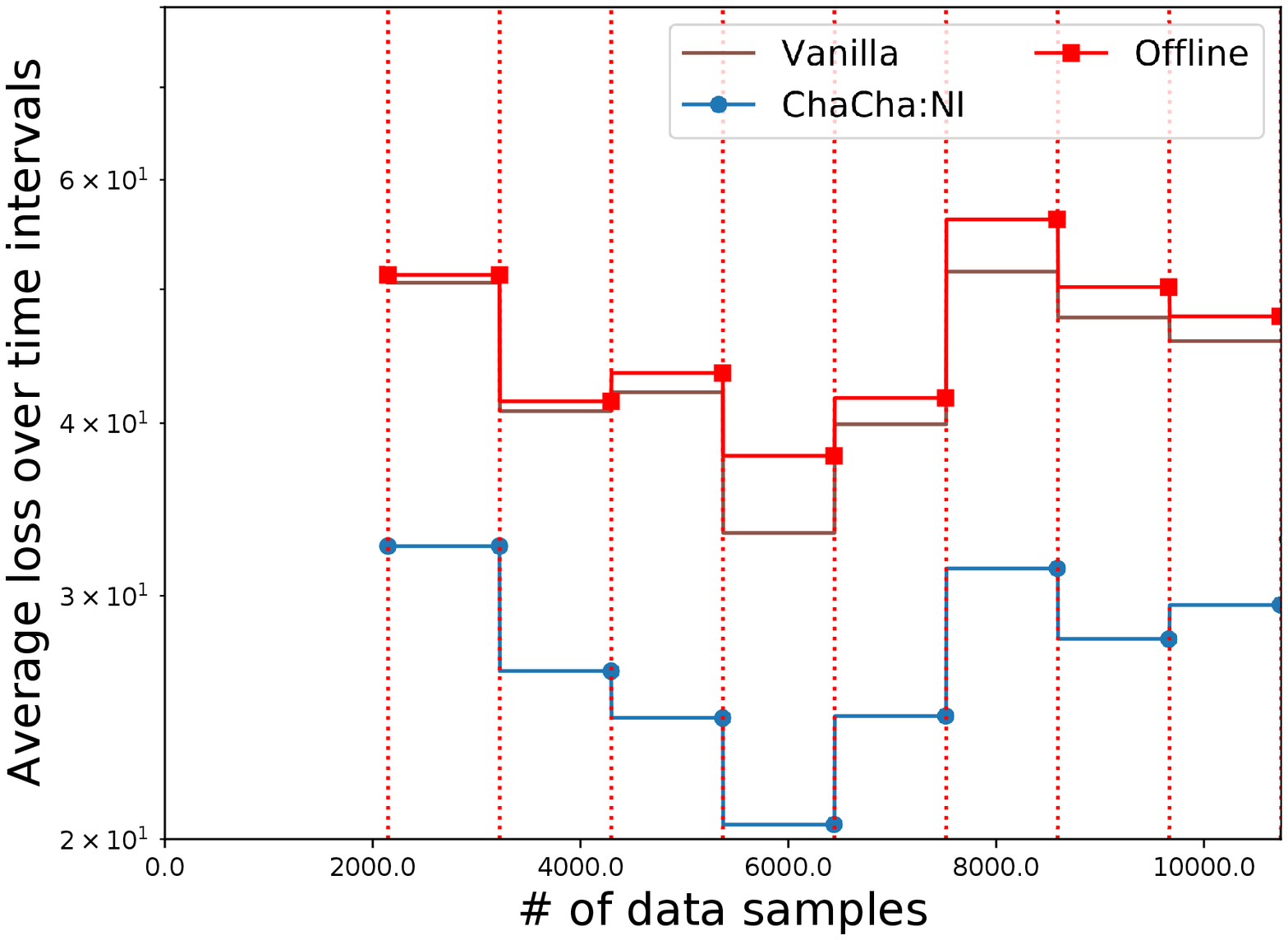}%
\end{subfigure}\hfill%

\caption{The existence of concept drifts on dataset \#42183. The method ‘Offline’ uses the first 20\% of the data to do training and is then tested without further model updates. In the second figure, for each of the methods, we report the average loss over each time intervals, split by vertical lines, starting from 20\% data samples. It helps illustrate a clear existence of concept drift in this dataset and the behavior of the compared methods in such a non-stationary environment.}
\label{fig:concept_drift}
\end{figure*}

\begin{table*} [ht]
\caption{Normalized scores (mean $\pm$ standard deviation) reported in Figure~\ref{fig:agg_oml}(b) and Figure~\ref{fig:ablation}.} \label{tab:scores}
\centering
\begin{tabular}{r | l l l  l  }
\hline
Dataset id &Random:NI & ChaCha:NI & ChaCha-w/o-Champion:NI & ChaCha-AggressiveScheduling:NI \\ \hline
1191 & 0.59 $\pm$ 0.35 & \textbf{0.64} $\pm$ 0.22 & -335.79 $\pm$ 44.34 & 0.00 $\pm$ 0.00 \\ 
1199 & \textbf{0.72} $\pm$ 0.53 & 0.10 $\pm$ 0.10 & -39.37 $\pm$ 0.33 & 0.00 $\pm$ 0.00 \\ 
1203 & \textbf{0.40} $\pm$ 0.35 & 0.38 $\pm$ 0.46 & -0.35 $\pm$ 0.01 & 0.00 $\pm$ 0.01 \\ 
1206 & 0.98 $\pm$ 0.03 & \textbf{1.02} $\pm$ 0.09 & -2.38 $\pm$ 0.04 & 0.00 $\pm$ 0.00 \\ 
1575 & 0.15 $\pm$ 0.06 & 0.13 $\pm$ 0.08 & \textbf{1.02} $\pm$ 0.05 & 1.00 $\pm$ 0.08 \\ 
201 & \textbf{0.12} $\pm$ 0.15 & 0.11 $\pm$ 0.13 & -6.15 $\pm$ 0.02 & -0.00 $\pm$ 0.00 \\ 
215 & 0.38 $\pm$ 0.37 & \textbf{0.41} $\pm$ 0.34 & -0.30 $\pm$ 0.04 & 0.01 $\pm$ 0.02 \\ 
23515 & 0.17 $\pm$ 0.15 & \textbf{2.94} $\pm$ 0.27 & -17.33 $\pm$ 0.03 & 1.04 $\pm$ 0.01 \\ 
344 & \textbf{0.34} $\pm$ 0.25 & 0.34 $\pm$ 0.30 & -7.18 $\pm$ 1.26 & 0.00 $\pm$ 0.00 \\ 
41506 & 0.71 $\pm$ 0.18 & \textbf{0.80} $\pm$ 0.18 & -6.53 $\pm$ 0.73 & 0.10 $\pm$ 0.09 \\ 
42183 & 0.11 $\pm$ 0.21 & \textbf{2.04} $\pm$ 0.08 & 0.16 $\pm$ 0.05 & 0.05 $\pm$ 0.01 \\ 
42496 & \textbf{0.19} $\pm$ 0.41 & -0.04 $\pm$ 0.02 & -28.09 $\pm$ 2.68 & 0.00 $\pm$ 0.00 \\ 
42729 & \textbf{0.46} $\pm$ 0.31 & 0.43 $\pm$ 0.25 & -54.10 $\pm$ 0.04 & -0.00 $\pm$ 0.00 \\ 
5648 & 0.97 $\pm$ 0.05 & \textbf{1.09} $\pm$ 0.04 & -0.01 $\pm$ 0.19 & 1.06 $\pm$ 0.04 \\ 
564 & 0.19 $\pm$ 0.21 & \textbf{0.74} $\pm$ 0.41 & -213.08 $\pm$ 14.48 & 0.00 $\pm$ 0.00 \\    \hline
\end{tabular}
\end{table*}

\begin{table*} [ht]
\caption{Normalized scores (mean $\pm$ standard deviation) reported in Figure~\ref{fig:agg_oml}(b).} \label{tab:score_ns+lr}
\centering
\begin{tabular}{r | l l l  l  }
\hline
Dataset id &Random:NI+LR & ChaCha:NI+LR \\ \hline
1191 & \textbf{1.00} $\pm$ 0.00 & 1.00 $\pm$ 0.00 \\ 
1199 & 1.00 $\pm$ 0.00 & \textbf{1.01} $\pm$ 0.03 \\ 
1203 & \textbf{0.11} $\pm$ 0.22 & -0.00 $\pm$ 0.01 \\ 
1206 & 0.95 $\pm$ 0.06 & \textbf{1.25} $\pm$ 0.03 \\ 
1575 & 0.62 $\pm$ 0.00 & \textbf{2.08} $\pm$ 0.54 \\ 
201 & \textbf{1.00} $\pm$ 0.00 & \textbf{1.00} $\pm$ 0.00 \\ 
215 & 0.09 $\pm$ 0.19 & \textbf{0.29} $\pm$ 0.39 \\ 
23515 & 1.00 $\pm$ 0.00 & \textbf{1.06} $\pm$ 0.17 \\ 
344 & 1.00 $\pm$ 0.00 & \textbf{1.15} $\pm$ 0.08 \\ 
41506 & 0.86 $\pm$ 0.00 & \textbf{0.95} $\pm$ 0.10 \\ 
42183 & 0.08 $\pm$ 0.00 & \textbf{1.86} $\pm$ 0.40 \\ 
42496 & 0.24 $\pm$ 0.00 & \textbf{0.39} $\pm$ 0.30 \\ 
42729 & \textbf{0.27} $\pm$ 0.10 & \textbf{0.27} $\pm$ 0.10 \\ 
5648 & 0.56 $\pm$ 0.31 & \textbf{1.11} $\pm$ 0.03 \\ 
564 & \textbf{1.00} $\pm$ 0.00 & \textbf{1.00} $\pm$ 0.00 \\ 
\hline
\end{tabular}
\end{table*}

Despite the i.i.d assumption in theoretical analysis, we do not exclude the possibility of non-stationary environments in our empirical evaluation (we intentionally do not shuffle the dataset such that potential concept drifts in the original datasets are preserved). In Figure~\ref{fig:concept_drift}, we show the results on an example dataset where concept drift exists. The results indicate a clear existence of concept drift, and \texttt{ChaCha} is still maintain its performance advantage. This appealing property is partly because of the base online learning algorithm's capability to adjust to the concept drifts and partly because of \texttt{ChaCha}'s progressive way of Champion promotion with the help of the \ConfigOracle.

\newpage
\section{Proof details} \label{sec:appendix_theory}
\begin{proof}[Proof of Lemma~\ref{lemma:tests_implication}].

According to the definition of $\epsilon_{c,t}$,

(1) $\forall m \in [M], c \in \cS_m$, with probability at least $1-\delta$,
\begin{align}
    & L^*_{\cF_c} - L^*_{\cF_{\Champion_m}} + 2 \epsilon_{c, t} + 3  \epsilon_{{\Champion_m}, t}   \geq L^{PV}_{c,t} + \epsilon_{c, t} - L^{PV}_{{C_m},t}  + 2 \epsilon_{{\Champion_m}, t}  = \overline{L}_{c, t} -  \underline{L}_{{\Champion_m}, t} +  \epsilon_{{\Champion_m}, t}
\end{align}
The above inequality indicates that as long as  $ L^*_{\cF_c} - L^*_{\cF_{\Champion_m}} + 2 \epsilon_{c, t} + 3  \epsilon_{{\Champion_m}, t} < 0$, $  \overline{L}_{c, t} -  \underline{L}_{{\Champion_m}, t} + 2  \epsilon_{{\Champion_m}, t} < 0$ which means that $c$ can pass the \BetterThan\ test when compared with $C_m$ at time $t$ and concludes the proof for Claim 1. 

(2) For Claim 2. 

When the \BetterThan\ test is triggered at time $t=t_{m+1}$, we have, with probability at least $1-\delta$,
\begin{align}
   & L^*_{\cF_{\Champion_{m}}}- L^*_{\cF_{\Champion_{m+1}}}   \geq  \underline{L}_{{\Champion_{m}},t} - \overline{L}_{{\Champion_{m+1}},t} 
  >  \underline{L}_{{\Champion_{m}},t} - (\underline{L}_{{\Champion_{m}},{t}} - \epsilon_{{\Champion_m}, t}) = \epsilon_{{\Champion_m}, t}
\end{align}
in which the second inequality is guaranteed by the fact that \BetterThan\ test is positive.

(3) For Claim 3. $\underline{L}_{c, t} - \overline{L}_{{C_m}, t} \leq L^*_{\cF_c} - L_{\cF_{C_m}}^*$ holds with probability at least $1-\delta'$. If $L^*_{\cF_c} < L_{\cF_{C_m}}^* < 0$, then with probability at least $1-\delta$, $\underline{L}_{c, t} - \overline{L}_{{C_m}, t} < 0$ (not passing the \WorseThan\ test).
\end{proof}

\begin{proof}[Proof of Proposition~\ref{prop:champion_loss_gap}]

Without affecting the order of the cumulative regret (w.r.t. $T$), we prove the regret bound assuming $c^* \in \cS_0$ (for the case $c^*$ is added at a particular time point $t'$, we only need to add an additional constant regret term related to $t'$).

According to Claim 1 of  Lemma~\ref{lemma:tests_implication}, during a particular phase $m$, i.e. $t_m \leq t \leq t_{m+1}-1 $, $\forall \cF \in \cS_m$, $ L^*_{\cF_{\Champion_m}} - L^*_{\cF_c} \leq 2 \epsilon_{c, t} + 3  \epsilon_{{\Champion_m}, t} $ must hold, otherwise phase $m$ would have ended. Since $c^* \in \cS_m$, we have, $\forall m \in [M]$, $\sum_{t=t_m}^{t_{m+1}-1}  (L^*_{\cF_{\Champion_m}} - L^*_{\cF_{c^*}})  < \sum_{t=t_m}^{t_{m+1}-1} (2 \epsilon_{{c^*}, t} + 3  \epsilon_{{\Champion_m}, t})$.

To account for the union over phases, we replace $\delta$ in $\epsilon_{c,t}$ from Eq.~\eqref{eq:epsilon}  by $ \delta' \coloneqq \delta/M$, and replace $|\cS_t|$ by $\max_{m \in M} |\cS_m|$ i.e.,  $\epsilon_{c,t} =   \comp_{\cF_{c}} \log(
\max_{m \in [M]} \frac{M|D_{t,c}||\cS_{m}|}{\delta}
)|D_{t,c}|^{p-1})$. By union bound we have the following inequality holds with probability at least $1-\delta$,
\begin{align} \label{eq:champion_loss_gap_decomp}
\sum_{m=0}^{M} \sum_{t=t_m}^{t_{m+1}-1}  (L^*_{\cF_{\Champion_m}} - L^*_{\cF_{c^*}}) & \leq \sum_{m=0}^{M} \sum_{t=t_m}^{t_{m+1}-1} 2 \epsilon_{{c^*}, t} + \sum_{m=0}^{M} \sum_{t=t_m}^{t_{m+1}-1}  3  \epsilon_{{\Champion_m}, t}= \sum_{t=1}^{T} 2 \epsilon_{{c^*}, t} + \sum_{m=0}^{M} \sum_{t=t_m}^{t_{m+1}-1}  3  \epsilon_{{\Champion_m}, t} \\ \nonumber 
\end{align}


The successive doubling resource allocation strategy ensures 
$\sum_{t=1}^{T} 2 \epsilon_{{c^*}, t} = O\big( \comp_{\cF^*} \max_{m \in [M]} \frac{|\cS_m|}{b}  T^p \log (\frac{TM|\cS_m|}{\delta}) \big) $.

Since we always keep the champion of each phase `live',  $\max_{ t_m <t <t_{m+t}}|D_{t, \Champion_m}| \geq t_{m+1}-t_{m} = N_{m}$. Thus we have,
\begin{align} \label{eq:champion_phase_m_bound}
\sum_{t=t_m}^{t_{m+1}-1} 3 \epsilon_{\Champion_m,t} 
& \leq 3 \comp_{\cF_{\Champion_{m}}} \sum_{t=t_m}^{t_{m+1}-1} |D_{t, \Champion_m}|^{p-1} \log T \\ \nonumber
 &  = O\big(\comp_{\cF_{\Champion_{m}}} N_m^p \log (\frac{\max_{m \in [M]}TM|\cS_m|}{\delta}) \big) \\ \nonumber 
 & = O\big(\comp_{\cF_{\Champion_{m}}} N_m^p \log T + \comp_{\cF_{\Champion_{m}}} N_m^p \log (\max_{m \in [M]}|\cS_m|) \big)
\end{align}

Now we provide an upper bound on the value of $\sum_{m=1}^M N_m^p$.

By Claim 2 of  Lemma~\ref{lemma:tests_implication}, 
\begin{align} \label{eq:N_sum_upper_bound}
   & L^*_{\cF_{\Champion_0}} - L_{\cF_{c^*}}^* \geq  
   =  \sum_{m=0}^{M-1} L^*_{\cF_{\Champion_{m}}} -  L^*_{\cF_{\Champion_{m+1}}} > \sum_{m=0}^{M-1} \epsilon_{{\Champion_m}, {t_{m+1}}} >   \comp_{\cF_{\Champion_m}} \sum_{m=0}^{M-1} N_{m}^{p-1} \log ( \frac{N_m}{\delta})
\end{align}
Now we discuss the properties of $\{N_m\}_{m\in[M-1]}$. Since $\sum_{m=0}^{M-1} N_{m}^{p-1} \log ( \frac{N_m}{\delta})$ converges, $ \sum_{m=0}^{M-1} \frac{1}{N_{m}^{1-p}} = \sum_{m=0}^{M-1} N_{m}^{p-1}$ converges. Since $1-p <1$, we have $N_{m}\geq  \Omega (m^{\frac{1}{1-p}})$, otherwise $ \sum_{m=0}^{M-1} \frac{1}{N_{m}^{1-p}}$ diverges according to convergence properties of Hyperharmonic series.

Since $T = \sum_{m=0}^{M} N_m$, we have  $T > \sum_{m=0}^{M-1} N_m > \sum_{m=0}^{M-1} \Omega (m^{\frac{1}{1-p}}) > \Omega(M^{\frac{2-p}{1-p}})$, which indicates $M < O(T^{ \frac{1-p}{2-p}})$. 
\begin{align} \label{eq:sum_N_M_bound}
    & \sum_{m=0}^M N_m^p   \leq (M+1)^{1-p} T^p = O(T^{\frac{1}{2-p}} )
\end{align}
in which the second inequality is based on Jensen's inequality.
Substituting Eq.~\eqref{eq:sum_N_M_bound} into Eq.~\eqref{eq:champion_phase_m_bound} and Eq.~\eqref{eq:champion_loss_gap_decomp} concludes the proof.
\end{proof}

\begin{proof}[Proof of Theorem~\ref{thm:final_regret_bound}]

\begin{align} \label{eq:regret_decompostion}
  & \sum_{t=1}^{T} (L_{{\Incumbent_t},t} -L^*_{\cF_{c^*}})  \sum_{m=0}^{M} \sum_{t=t_m}^{t_{m+1}-1} (L_{{\Incumbent_t},t} -  L^*_{\Champion_m} ) + \sum_{m=0}^{M}  \sum_{t=t_m}^{t_{m+1}-1} ( L^*_{\Champion_m} - L^*_{\cF_{c^*}})\\ \nonumber
\end{align}

The second term in the right-hand side of Eq.~\eqref{eq:regret_decompostion} can be upper bounded by Proposition~\ref{prop:champion_loss_gap}. Now we upper bound the first term of the right-hand side of Eq.~\eqref{eq:regret_decompostion}. $\forall m \in \{0, 1, \cdots, M\}$,

\begin{align}
\sum_{t=t_m}^{t_{m+1}-1} (L_{{\Incumbent_t}, t} -  L^*_{\Champion_m}) 
& \leq  \sum_{t=t_m}^{t_{m+1}-1} \epsilon_{{\Incumbent_t}, t} + L^{PV}_{{\Incumbent_t}, t} - L^{PV}_{{\Champion_m}, t} + \epsilon_{{\Champion_m}, t} \\ \nonumber
& \leq  \sum_{t=t_m}^{t_{m+1}-1} \epsilon_{{\Champion_m}, t} +  L^{PV}_{{\Champion_m}, t}  - L^{PV}_{{\Champion_m}, t} + \epsilon_{{\Champion_m}, t}  \\ \nonumber
& =  \sum_{t=t_m}^{t_{m+1}-1} 2\epsilon_{{\Champion_m}, t} = O(\comp_{\cF_{\Champion_m}} N_m^p \log T)
\end{align}
in which the last inequality is based on the fact that ${\Incumbent_t} = \argmin_{{c} \in \cB_t} ( L^{PV}_{{c}, t} +  \epsilon_{{c}, t})$.  Substituting conclusion in Eq~\eqref{eq:sum_N_M_bound} which is proved in Proposition~\eqref{prop:champion_loss_gap}, into the above inequality, we have
\begin{align} \label{eq:regret_comparing_to_champion}
    & \sum_{m=0}^{M} \sum_{t=t_m}^{t_{m+1}-1} (L_{{\Incumbent_t}, t} - L^*_{\Champion_m}) = O(  \max_{m \in [M]} \comp_{\cF_{\Champion_m}}  T^{\frac{1}{2-p}} \log T)
\end{align}

Substituting Eq.~\eqref{eq:regret_comparing_to_champion} and the conclusion in Proposition~\ref{prop:champion_loss_gap} into Eq.~\eqref{eq:regret_decompostion} finishes the proof.
\end{proof}

\end{document}